# Explanation of Probabilistic Inference for Decision Support Systems

Christopher Elsaesser
MITRE Washington Artificial Intelligence Center and
Carnegie-Mellon University Department of Engineering & Public Policy

**Abstract**
This paper reports work in progress on an explanation facility for Bayesian conditioning aimed at improving user acceptance of probability-based decision support systems. Design of the facility, which appears to be reasonably domain-independent, is based on an information processing model that accounts both for biased and normative behavior in reasoning about conditional evidence. Preliminary results indicate that the facility is both acceptable to naive users and effective in improving understanding of Bayesian conditioning.

**Introduction**

Decision aids appear necessary to overcome the systematic errors to which people are prone in uncertainty reasoning (Kahneman and Tversky, 1982) but the availability of automated tools has not led to significant reliance on normative techniques of reasoning. In fact automating normative processes seems to be detrimental to their introduction in many applications. Acceptance problems may stem from the translation between mental models of a decision situation to a formal model and back from the implications of the formal model to terms users can understand.

Because rule-based expert systems are decision aids based on descriptive models of human problem solving behavior, the assumption is often made that a recapitulation of program actions paraphrased in natural language will automatically be understandable to users. When rule-based inference was extended in MYCIN and Prospector to account for uncertainty, the typical rule backchaining technique of explanation was extended in a straight-forward manner. Here is a MYCIN rule as it would be printed in an explanation:

IF:   (1) The identity of the organism is not known with certainty, and
      (2) The stain of the organism is gramneg, and
      (3) The morphology of the organism is rod, and
      (4) The aerobicity of the organism is aerobic
THEN:     There is strongly suggestive evidence (.8) that the class of the organism is enterobacteriaceae.

But the explanation facilities of most expert systems do not explain the updating of certainty that results from a rule executing.

There recently have been efforts to develop explanation facilities for decision support systems which employ normative methods (Spiegelhalter, 1985; Reggia and Perricone, 1985; Politser, 1984; Norton, 1986; Langlotz, Shortliffe, and Fagan, 1986). However, none of the approaches is based on an explicit model of cognition that accounts for the dissonance that occurs when user's compare normatively calculated probability with an internally generated estimate. In fact, the general belief in the AI community is that representation of uncertainty based on probability is not explainable:

> It is difficult to imagine what explanation of its actions the program [based on Bayes' rule] could give if it were queried about computed probabilities. No matter what level of detail is chosen, such a program's actions are not (nor were they intended to be) a model of the reasoning process typically employed by physicians. Although they may be an effective way for the computer to solve diagnosis problems, there is no easy way to interpret these actions in terms that will make them comprehensible to humans unacquainted with the program. (Barr and Feigenbaum, 1982)

The current approach to building understandable AI systems that reason with uncertainty is to develop uncertainty mechanisms that are *descriptive* of human reasoning, eschewing normative probability. While advocates claim these alternative calculi address the problem of cognitive dissonance, the tendency to deviate from normative rules is unacceptable for many applications. In any case, claims that these alternatives are descriptive of human reasoning have not been verified.

## Understanding and Explanation of Bayesian Inference

The explanation technique described herein is based on the theory that Bayesian inference is, in fact, *descriptive* of some human reasoning. A process model of uncertainty reasoning is proposed that (1) accounts for the systematic biases that people exhibit in assessing probability due to neglect of prior probabilities (Kahneman, Slovic, and Tversky, 1982), (2) includes procedures that are consistent with Bayes' rule, and (3) accounts for the selection of one set of procedures over the other.[1]

*Heuristics and Biases*

The heuristics of representativeness, availability, and anchoring and adjustment are procedures that account for errors in assessing probabilities due to neglect of base rates (Kahneman and Tversky, 1982). Consider a screening test known to detect the presence of drugs in 90% of users tested. An error due to neglect of base rates would be the inference that the probability of drug use by a person who tests positive is 0.9; *i.e.*, the sensitivity of the test is mistaken for the posterior probability. This behavior is accounted for in my process model by a production which produces the heuristic that people who test positively are *representative* of people the test was designed to detect:

IF a *?class* has *?feature* with probability *?p* and
  *?individual* has *?feature*
THEN *?individual* is a *?class* with probability *?p*

When *?class*, *?feature*, *?p*, and *?individual* are instantiated with elements in working memory, the resulting inference would be:

IF a *drug user* has *positive test* with probability *0.9* and
  *Joe Employee* has *positive test*
THEN *Joe Employee* is a *drug user* with probability *0.9*[2]

---

[1] I express the model in the ACT architecture of cognition (Anderson, 1983). In ACT, productions in procedural memory are selected for processing based on information in internal Short Term Memory along with salient information in foveal vision.

[2] Note that the MYCIN rule in the preceding section is essentially of the same form. The (.8) in the rule is the certainty *associated with the rule*, not the resulting certainty of the hypothesis. Thus, such explanations will reinforce disregard of base rate information.

395

Similarly, the heuristics of availability and anchoring and adjustment can be accounted for under the assumptions of ACT.

*Bayesian Thinking in Humans*

There is evidence suggesting that productions that account for prior probability are part of people's procedural knowledge. For example, Kahneman and Tversky (1982) found that people use prior probabilities correctly when they have no other information but effectively ignore them when a description is introduced, even when the description is totally uninformative. Fischhoff, Slovic, and Lichtenstein (1979) found that subjects become sensitive to prior probabilities and to the reliability of evidence when successive problems are encountered that differ only in these variables; biases appear when the context of the problem has not been seen often enough to allow these salient features to be recognized.

Including productions that account for these observations in a process model is not inconsistent with the existence of heuristics such as representativeness described above. A production that assigns posterior belief to be the prior probability is sufficient when only prior information is available. Processing descriptive information could reduce the salience of prior information, resulting in the representativeness heuristic being applied inappropriately.[1] But extending this process model with productions that properly account for prior probability requires the identification of some additional productions that plausibly could be part of basic inductive knowledge. Such productions are suggested by Polya's "shaded inductive patterns":

| A implies B | A implies B |
| B almost certain anyway | B without A hardly credible |
| B true | B true |
| A very little more credible | A very much more credible |

Their intuitive appeal indicates these "productions" may be part of basic inductive knowledge. They seem consistent with the apparent relationship between surprise and modification of belief; unexpected observations revise the likelihood of a hypotheses more than evidence consistent with current belief:

> Perhaps the main biological function of surprise is to jar us into reconsidering the validity of some hypotheses ... To be surprised ... is to obtain a substantial Bayes factor  (Good, in Shafer, 1982, p. 343)

Importantly, Polya's patterns describe reasoning that can be modeled by Bayes' rule. For example, if "B almost certain anyway" is interpreted as the marginal $P(B)$ (the divisor in Bayes' rule) being close to 1.0 and "A implies B" means $P(B \mid A)$ is close to 1.0, then by Bayes rule the probability of A would be "very little more" than what it was prior to observing B.[2]

The purpose of an explanation facility based on this model is to elicit the use of appropriate procedure's by the user. The critical factor is accounting for how alternative procedures are selected.

*Linguistic Representation of Probability*

Most investigations of subjective inference have been concerned with the underlying procedures used to process information rather than with modes of representation. But the representation of information in working memory can influence which problem solving procedures are selected (Simon, 1979). If humans have a variety of computational procedures for reasoning under uncertainty then improved performance in inference tasks should be observed when more suitable representations are used.

---

[1] Salience of priors is largely under the control of a decision support system since, according to ACT, working memory includes elements in foveal vision.

[2] Exactly how much more depends on the prior.

Cohen and Grinberg (1983) suggest that numerical representations of uncertainty detract from understanding and that human judgment may be better than it appears from numerically expressed estimates. Linguistic expression of uncertainty might require a less complicated transformation for storage in short term memory, introducing less distortion than a numerical representation and resulting in better understanding (Freksa, 1982). Zimmer (1983) found that subjects judged probabilities better when verbal modes were employed and Zimmer (1985) reports evidence that verbal processing of knowledge of uncertainty reduces biases of conservatism and negligence of regression. His results indicate that people are much closer to optimal Bayesian revision when they are allowed to use linguistic expressions of probability. Oden (1977) obtained similar results for conjunctive and disjunctive reasoning:

> [probability rules] provide a substantially better fit to the data [than the rules of Possibility theory] for every (data) matrix and for the great majority of the subjects. (Oden, 1977)

The observations that linguistic representation of uncertainty appear to improve reasoning support the conjecture that alternative processes for uncertainty exist and that representation is a determining factor in which processes are selected. It follows that stating explanations in linguistic form using relevant inference procedures will improve understanding.

### An Explanation Facility for Bayesian Inference

An explanation facility consisting of templates which produce linguistic representations of Bayesian inference has been developed. The templates are based on the "shaded inductive patterns" of Polya (1954) and natural language term sets expressing probabilities and changes in probabilities. The goal was to make the facility as domain independent as possible. This section describes some details of a prototype implementation with emphasis on the psychological basis for implementation decisions. The first step was to choose a representative set of terms.

*Lexicon of Probability*[1]

For maximum explanatory power, translations from probabilities to natural language should convey as much information as possible but no more than the capability of the user to distinguish differences (to minimize cognitive load). The ability to distinguish meaningful differences in symbolic information is assumed to be governed by the capacity of Short Term Memory; about 5 symbols (Simon, 1985). Note that in linguistic form, probabilities tend to have a natural subdivision point; hypothesis of probability less than the subdivision point are referred to as "unlikely", hypothesis with probability greater than the subdivision point are "likely". Thus, granularity of the lexicon can be substantially decreased by assuming that "chunking" occurs at the subdividing point. If the ability to easily distinguish levels of probability below and above the natural subdivision point are assumed to be governed by the hypothetical

---

[1] It is clear that the lexicon of probabilities will be domain dependant and require calibration.

397

capacity of STM, minimum granularity (maximum list size) for an effective lexicon of probability appears to be 11 (5 phrases on either side of a subdivision point term). Based on this analysis, the term set shown in Table 1 was adapted from Lichtenstein and Newman (1967) for use in the prototype.[1]

Table 1: Term set for expressing probabilities

| TERM | RANGE OF APPLICATION[2] |
|---|---|
| Almost certain | 0.99 - 0.91 |
| Highly probable | 0.90 - 0.82 |
| Quite likely | 0.81 - 0.73 |
| Rather likely | 0.72 - 0.64 |
| Better than even | 0.63 - 0.55 |
| Fair chance | 0.54 - 0.46 |
| Not quite even chance | 0.45 - 0.37 |
| Somewhat unlikely | 0.36 - 0.28 |
| Rather unlikely | 0.27 - 0.19 |
| Improbable | 0.18 - 0.09 |
| Highly improbable | 0.08 - 0.01 |

The lexicon selected to express the *change* from prior to posterior probability is more important than the lexicon of probabilities because change in belief is what Bayesian conditioning models. In contrast to the lexicon of probability just discussed, there is no obvious "cognitive break point" in comparison terms. Thus, the practical upper limit on term set size should be the same as the assumed symbol capacity of short term memory (*i.e.*, 5). However, two lists are required, one for decreasing likelihood (disconfirming evidence) and one for increasing likelihood (confirming evidence). It remains an issue for further research to determine the best (most consistently used, *etc.*) set of terms. Table 2 shows the lists used in the prototype.

Table 2: Term set for expressing change in probabilities

| Decreasing Certainty | Increasing Certainty |
|---|---|
| a great deal less likely | a great deal more likely |
| much less likely | much more likely |
| quite a bit less likely | quite a bit more likely |
| somewhat less likely | somewhat more likely |
| a bit less likely | a bit more likely |

A major issue still under investigation is the exact nature of the mapping from the cross product space of prior and posterior probabilities to phrases describing change in belief. Consider describing a change in probability from 0.91 to 0.95 with a change from 0.01 to 0.05. One might say that 0.95 is "slightly more likely" than 0.91, but that 0.05 is "a great deal more likely" than 0.01, even though the absolute differences are the same.

---

[1] Initial development assumed the range of probabilities to be [0,1] and the subdivision point to be 0.5. These assumptions are domain dependent.

[2] The midpoints of term ranges should be equally spaced for language pragmatics when probabilities of interest are distributed over [0,1] (Zimmer, 1985). Where probabilities tend to cluster at one end of the unit interval, some transformation appears to be necessary. The characteristics of that transformation are not addressed in this paper.

Oden (1979) found that subjects were consistent in their comparison of relative belief, reporting experimental evidence supporting the following relation:

$$\text{Rel}(A, B) = \frac{\text{Truth}(A)}{\text{Truth}(A) + \text{Truth}(B)}$$

where Rel(A, B) is the relative truthfulness of statement A in relation to statement B and Truth(x) is a linguistic expression of belief in proposition x. The prototype was implemented using this relation; its domain independence remains an issue for further investigation.

*Templates*

Polya's Patterns of Plausible Inference are the inspiration for English language sentences in which these words are embedded. The patterns can easily be rephrased into production-like sentence structures which experience in deductive expert systems has shown to be acceptable to users.

The marginal probability of evidence is the key used for selection of the templates. For example, a "small" marginal indicates that the evidence is unexpected or surprising. Because such a result strongly disconfirms prior belief, substantial belief revision is necessary. Conversely, a "large" marginal indicates evidence confirms prior belief; belief revision will be minimal.

With the marginal as selection criteria for templates, only three basic situations will be encountered corresponding to confirming evidence, disconfirming evidence, and a middle ground where prior belief accommodates either result. When only a single point of reference is used in an application, the probability at which observing evidence is neither "likely" nor "unlikely" occurs when the marginal is in a neighborhood of 0.5; a point which appears to be application and domain independent. In the prototype sentences were structured so that operational meaning is conveyed by words selected from term sets, making only two templates necessary. In hindsight, a single template appears sufficient for this application. Figures 1 and 2 show examples generated by the facility.

Figure 1: Explanation of effect of anticipated evidence.[1]

<u>Based only on its structure</u>, it is *not quite an even chance* that <u>P345-22 is a carcinogen</u>.

Because a <u>Positive Sister-Chromatid Exchange test</u> is *quite likely* for a <u>pyrrolizidine</u>, the hypothesis that <u>P345-22 is a carcinogen</u> is *somewhat more likely,* making it *highly probable* that <u>P345-22 is a carcinogen.</u>

---

[1]*Italics* indicates the words selected from the term sets previously discussed. <u>Underlining</u> indicates application dependent terms.

**Figure 2:** Explanation when evidence is neither anticipated nor surprising

<u>Based only on its structure</u>, it is *somewhat unlikely* that <u>P98-21 is a carcinogen.</u>

There is a *fair chance* of a <u>Negative L5178Y test</u> for a <u>benz-a-anthracene</u>, making it *a bit less likely* that <u>P98-21 is a carcinogen.</u> It is *improbable* that <u>P98-21 is a carcinogen</u>.

### Acceptability

The preceding sections relate a theory of understanding that provide the basis for explanation of Bayesian conditioning. The usual AI technique for validating the acceptability of such a program is to demonstrate it for a target population of users. The testimony of potential users is, of course, the weakest of acceptability tests, but is prudent before continuing development. To that end, a decision support application in which normative inference was a central issue to the system's acceptance was selected to prototype this explanation technique. The Environmental Protection Agency's (EPA) risk analysis of new chemicals mandated by the Toxic Substances Control Act provided an appropriate domain. A prototype decision support system for analysis of short-term mutagenicity test results was designed. The purpose of the system is to determine the probability of carcinogenicity of a new chemical. Based on a chemical's structure, a prior probability is assigned. The program calculates the posterior probability of carcinogenicity based on test results and expert knowledge of the applicability of the tests to various chemical classes. If requested by the user, the most definitive follow-up test is recommended. Both the initial analysis and the recommendation of the follow-up test are explained using the facility described herein.

An Agency consultant familiar with EPA decision makers cautioned that they would "never use" a Bayesian decision support system because they do not understand the probabilities (Clifford, 1985). But based on interviews with EPA toxicologists, it appears the facility produces understandable and acceptable explanations of single step Bayesian conditioning.

### Effectiveness

Such a demonstration hardly constitutes validation of the model or supports the contention that Bayesian conditioning can be made understandable enough to remove it as an impediment to acceptability of normative decision support systems. The central issue is whether decision makers "understand" output enough to act on the recommendations of the program. This question is obviously difficult to test directly; it was necessary to construct an experiment that tested a side effect.

It is easy to demonstrate that subjects *do not* understand Bayesian conditioning. Present subjects with a scenario which includes information about prior probability (*e.g.*, a population average) and some evidence that is conditionally related to the target condition (*e.g.*, an indicator which exhibits false negative and false positive rates) and most subjects will ignore the prior, mistaking the true positive rate to be the posterior probability. When informed of the normative posterior the subject will be confused how such a result could be correct; *i.e.*, he will not "understand" it. If, in the (unannounced) presence of an explanation, a subject's answer is consistent with the normative posterior we might conclude he "understands" the inference. This is the essential outline of the experiment conducted to determine the effectiveness of the explanation facility.

*Design*

Seven problems were presented in random order to each subject. The problems were duplicities of, or isomorphic to, inference problems from the "heuristics and biases" literature. For each problem the subject selected one of five sub-intervals of the unit interval in which he thought the probability in question fell. A between-groups design was used. A control group was given the original problems and the treatment group was given identical materials with an explanation generated by the prototype facility. Unlike a decision support system, there was no attention drawn to the explanation; it was simply included in the text immediately prior to the question concerning the probability.

*Subjects*

Two groups of approximately 50 subjects each returned materials. The groups were randomly selected from a list of Members of the Technical Staff of the MITRE Corporation Washington Center. Follow-up questions indicated that the two groups were essentially the same in all relevant respects (typical level of education, exposure to decision analysis concepts, *etc.*).

*Results*

Replicating results from the literature, the majority of subjects in the control group neglected prior probability and chose the "biased" answer. As the histograms of subject responses in the figures below demonstrate, exposure to the explanation resulted in a discernable shift to the interval containing the normative solution.

While this experiment only indirectly addresses the central issues of this research, it does provide encouragement to continue development of both the model and the facility.

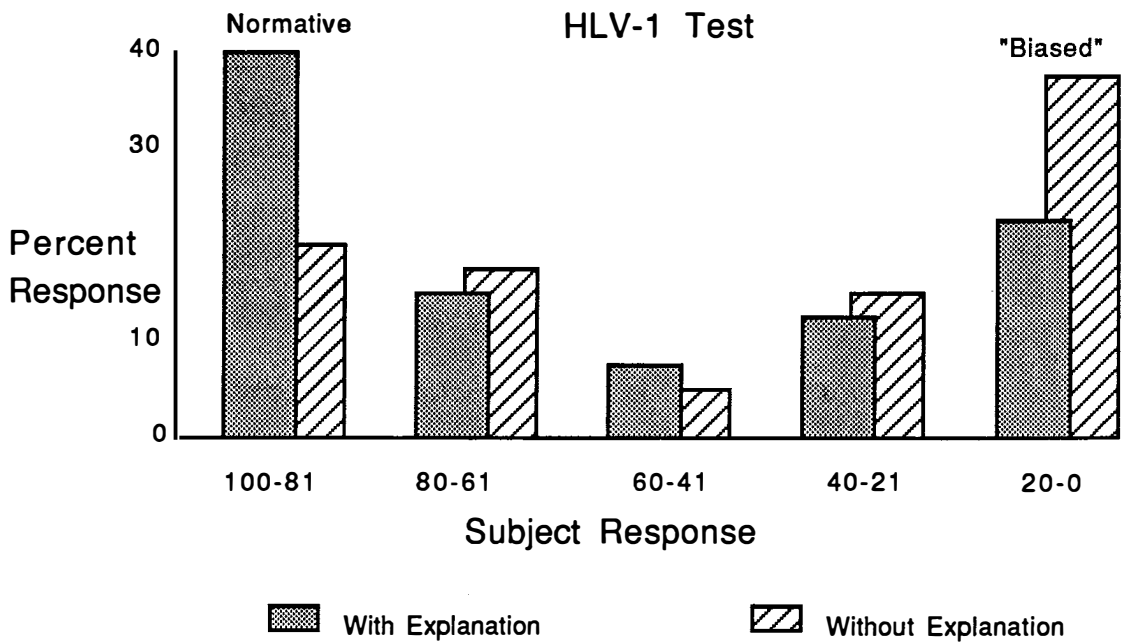

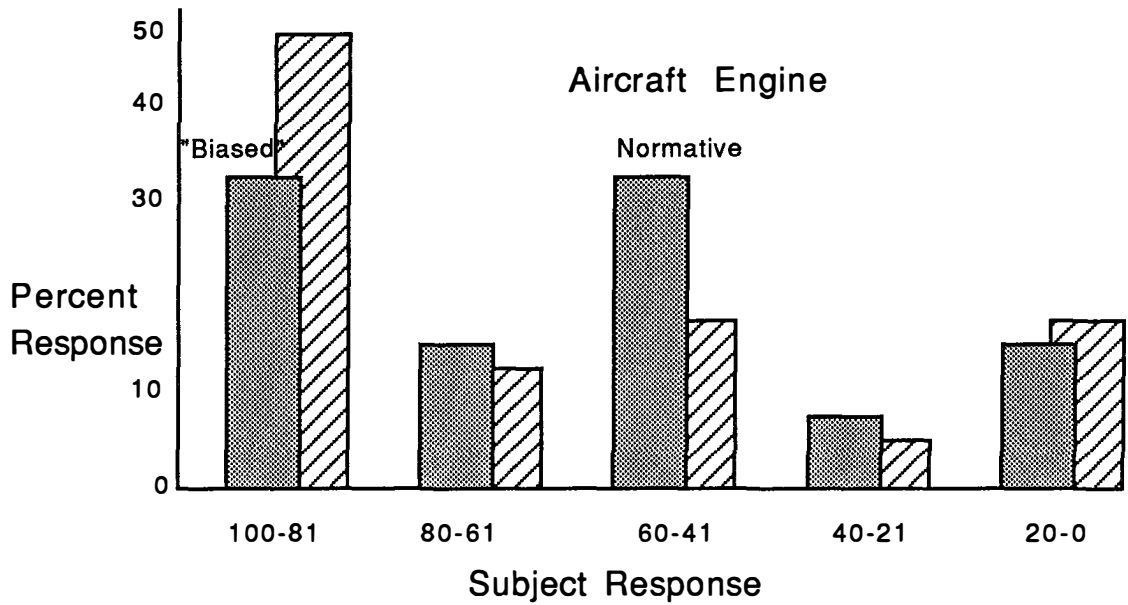

Subject Response

**Summary**

This paper describes an explanation facility based on a cognitive model which predicts that conditional reasoning can be manipulated to produce results described by normative rules of probability. The maximum effective length of the lexicons of probability, the transformation from the cross product space of prior and posterior probability to the term set used to describe change in probability, and the basic sentence structures are suspected to be domain-independent features of the explanation technique. Preliminary evidence indicates that the facility is both acceptable to potential users and effective in overcoming the "haze of Bayes" (Polister, 1984).

**References**

Anderson, John R. The Architecture of Cognition. Harvard University Press, Cambridge, Mass. 1983.

Barr, A., Feigenbaum, E. The handbook of Artificial Intelligence, Vol 1. William Kaufmann, Inc. Los Altos, California 1981.

Clifford, Paul. Personal communication, July, 1985.

Cohen, Paul R., Grinberg, Milton R. A Theory of Heuristic Reasoning About Uncertainty. The AI Magazine, p. 17-24, Summer 1983.

Freksa, C. Linguistic description of human judgments in expert systems and in The 'soft' sciences. In Gupta & Sanchez (82) p. 297-305

Kahneman, Daniel, Slovic, Paul, and Tversky, Amos. Judgment under uncertainty: Heuristics and biases. Cambridge University Press, 1982.

Kahneman, D., and Tversky, A. On the study of statistical intuitions in Kahneman, Slovic, and Tversky (1982).

Langlotz, C.P., Shortliffe, E.H., Fagan, L.M. A Methodology for Computer-Based Explanation of Decision Analysis, Medical Computer Science Group Working Paper KSL-86-57, November, 1986.

Lichtenstein, S., Newman, J.R. Empirical scaling of common verbal phrases associated with numerical probabilities, in Psychon. Sci., 1967, Vol. 9 (10).

Norton, S., An Explanation Mechanism For Bayesian Inferencing Systems, in Proceedings of the Workshop on Uncertainty in Artificial Intelligence , Radio Corporation of America and the American Association for Artificial Intelligence, Philadelphia, Pa, 1986, pp. 193-200.

Oden, G.C., Integration of fuzzy logical information, Journal of Experimental Psychology: Human Perception and Performance. vol. 3, no. 4, (1977), 565-575.

Polya, George. Mathematics and Plausible Reasoning, Vol. II: Patterns of Plausible Inference. Princeton University Press, 1954.

Politser, P. E. Explanations of statistical concepts: Can they penetrate the Haze of Bayes? Methods of Information in Medicine 2 (1984) p.99-108.

Quinlan, J.R. Consistency and Plausible Reasoning. Proceedings of the Eighth International Joint Conference on Artificial Intelligence, William Kaufmann, Inc. Los Altos, Calf. 1983 p. 137-144

Reggia, James A., and Perricone, Barry T. Answer justification in medical decision support systems based on Bayesian classification. Comput. Biol. Med. Vol. 15, No. 4, 1985, pp. 161-167.

Shafer, Glenn. Lindley's Paradox, JASA Vol. 77, No. 378, June 1982 pp. 325-351

Simon, H. Personal communication, October, 1985.

Simon, H. Models of Thought. Yale University Press, New Haven, Conn. 1979.

Spiegelhalter, David J. A statistical view of uncertainty in expert systems, Presented at workshop on AI and statistics, Bell Telephone Laboratories, April 11-12th, 1985

Zimmer, Alf C. The Estimation of Subjective Probabilities via Categorical Judgments of Uncertainty. Proceedings of the Workshop on Uncertainty and Probability in Artificial Intelligence, UCLA August 14-16, 1985, p. 217-224.

Zimmer, Alf C. Verbal vs. numerical processing of subjective probabilities. in Decision Making Under Uncertainty, R.W. Scholz (editor), Elsevier Science Publishers B.V. (North-Holland), 1983, p. 159-182.